\documentclass{article}
\usepackage{ijcai22}

\usepackage{times}
\usepackage{soul}
\usepackage{url}
\usepackage[hidelinks]{hyperref}
\usepackage[utf8]{inputenc}
\usepackage[small]{caption}
\usepackage{graphicx}
\usepackage{amsmath}
\usepackage{amsthm}
\usepackage{booktabs}
\usepackage{algorithm}
\usepackage{algorithmic}
\usepackage{multirow}
\usepackage{multicol}
\usepackage{amssymb}

\title{Towards Fewer Labels: Support Pair Active Learning for Person Re-identification}

\author{
Paper ID: 836
}

\author{
Dapeng Jin
\and
Minxian Li
\affiliations
Nanjing University of Science and Technology\\
\emails
\{jindapeng, minxianli\}@njust.edu.cn
}

\begin{document}

\maketitle

\begin{abstract}
Supervised-learning based person re-identification (re-id) require a large amount of manual labeled data, which is not applicable in practical re-id deployment.
In this work, we propose a Support Pair Active Learning (SPAL) framework to lower the manual labeling cost for large-scale person re-identification.
The support pairs can provide the most informative relationships and support the discriminative feature learning.
Specifically, we firstly design a dual uncertainty selection strategy to iteratively discover support pairs and require human annotations.
Afterwards, we introduce a constrained clustering algorithm to propagate the relationships of labeled support pairs to other unlabeled samples.
Moreover, a hybrid learning strategy consisting of an unsupervised contrastive loss and a supervised support pair loss is proposed to learn the discriminative re-id feature representation.
The proposed overall framework can effectively lower the labeling cost by mining and leveraging the critical support pairs.
Extensive experiments demonstrate the superiority of the proposed method over state-of-the-art active learning methods on large-scale person re-id benchmarks.
\end{abstract}

\section{Introduction}
\label{sec:intro}
Person re-identification aims to identify the same person across non-overlapping camera views, which is a challenging task in computer vision.
In recent years, with the increasing demand for public safety and the increasing number of surveillance cameras, person re-identification has attracted extensive research.
The supervised person re-id approaches achieve promising performance, relying on a large amount of manually labeled IDs for each person images.
However, these approaches are not applicable in large-scale person re-identification deployment.

For reducing the annotation costs, three categories of person re-identification method are proposed.
(1)Unsupervised learning methods \cite{lin2019aBottom,lin2020unsupervised,wang2020unsupervised,ge2020selfpaced} do not require human labeled ID annotations, and train the model on unlabeled images. However, without cross-view pairwise ID labels, this kind of methods cannot learn the discriminative feature representation for the significant appearance change across camera views.
(2)Semi-supervised learning methods \cite{wu2018exploit,li2019unsupervised} need ID information guided data (such as one image per ID) to start the model training. But actually, the labeling cost of ID information is exhaustively high, which equals to the labeling cost in supervised learning methods.
(3)Active learning methods \cite{martinel2016temporal,liu2017early,liu2019deep,hu2021cluster} aim to select a subset of the most representative image pairs for labeling and training. Due to the iterative labeling procedure, the manual annotation cost of active learning is limited and controllable.

Although existing active learning methods can balance the re-id performance and the labeling cost, the labeling amount of pairs is still quite high.
We observe that, the relations of pairwise data can be predicted accurately by the unsupervised learning framework, e.g. \cite{ge2020selfpaced}.
Inspired by this, we consider an active learning method that focuses on mining some critical pairs which are derived from the clustering structure.
If these critical pairs are labeled, we expect they can effectively and efficiently help re-id model training.

For this purpose, we consider a Support Pair Active Learning (SPAL) framework for person re-identification.
To reduce the labeling cost, the proposed method focuses on mining and leveraging \textit{support pairs} from unlabeled images (no ID labels).
The \textit{support pairs} refer to the critical pairwise data which can provide the most informative relationships and support the discriminative feature learning.
Specifically, two kinds of support pairs: uncertain positive pair and uncertain negative pair are selected by a dual uncertainty selection strategy and then required to be labeled by experts.
To extend the relationship of the labeled support pairs, we propose a constrained clustering algorithm to achieve a reliable clustering by meeting both must-link and cannot-link constraints derived by the labeled support pairs.
For effectively optimizing the feature representation model, a hybrid learning loss is proposed, consisting of an unsupervised contrastive loss and a supervised support pair loss.
The SPAL approach gradually achieves both a reliable clustering and a discriminative feature representation model.
By support pairs selection and effective learning strategy, the SPAL approach attains satisfying re-id performance with low manual labeling cost.

Our contributions are as follows:
(1) We formulate a Support Pair Active Learning (SPAL) framework for large-scale person re-identification. 
Compared with supervised learning methods, this model can achieve competitive re-identification performance, but with much lower labeling cost.
(2) We propose a dual uncertainty selection strategy to select the most informative sample pairs, termed support pairs, for the human annotations.
(3) We present a constrained clustering algorithm by utilizing the labeled support pair to achieve a reliable clustering, and propose a hybrid learning loss, consisting of an unsupervised contrastive loss and a supervised support pair loss, to effectively optimize the feature representation model.
(4) Extensive comparative experiments demonstrate the two-fold advantages of SPAL: high re-id performance and low labeling cost on four large-scale person re-id benchmarks: Market-1501\cite{zheng2015scalable}, DukeMTMC-ReID\cite{zheng2017unlabeled},  MSMT17\cite{wei2018person}, and LaST\cite{shu2021large}.

\section{Related Work}
\label{sec:related}


\noindent{\textbf{Unsupervised Learning Person Re-ID.}}
Depending on whether extra data is used in training stage, unsupervised learning methods can be devided into two categories:
(1) Pure unsupervised learning.
Most unsupervised learning re-id methods
\cite{lin2019aBottom,zeng2020hierarchical} 
adopted a clustering algorithm to produce pseudo labels for each cluster, and then train re-id model by pseudo labels.
(2) Unsupervised domain adaptation (UDA).
UDA approaches aim to transfer the learned knowledge from a labeled source domain to an unlabeled target domain.
Specifically, this kind of method can be grouped as pre-trained model, synthesis model and joint-learning model.
The methods can be classified into three groups:
Source domain pre-trained methods \cite{fu2019self,yang2020asymmetric}, 
Image-synthesis based methods \cite{wei2018person,tang2020cgan}, 
and Joint-learning based methods \cite{zhong2019invariance,ge2020selfpaced}. 
Although no labeling cost for unsupervised learning approaches, it has inferior performance on large-scale person re-identification due to the lack of pairwise annotations. 
Compared with unsupervised re-id methods, we manage to discover and annotate a few of pairwise samples to promote the re-id model's ability.

\noindent{\textbf{Semi-Supervised Learning Person Re-ID.}}
Semi-supervised learning methods \cite{wu2018exploit,li2019unsupervised} 
train re-id model both on pre-labeled and unlabeled samples.
Basically, most of these methods work on one-shot learning setting, which need pre-label one sample per each ID. 
For example,
Wu et al. \cite{wu2018exploit} initialized a CNN model using pre-labeled data per ID, and then adopted a step-wise learning approach to update the CNN model.
Li et al. \cite{li2019unsupervised} used pre-labeled within-camera tracklet per ID to initialize a deep model, and then incrementally discover cross-camera tracklet association to improve the representation capability of deep model.
Although semi-supervised learning methods improve re-id performance, the one-shot labeling strategy is actually not practical for re-id task. The true labeling cost equals to supervised re-id methods, because it need ID information. 
Different from the impractical labeling strategy of semi-supervised re-id, SPAL entirely starts from unlabeled samples (no ID label), and gradually mines and labels a small number of critical pairs.

\noindent{\textbf{Active Learning Person Re-ID.}}
Active learning re-id methods aim to select a subset of pairwise data to label in the training stage. 
Liu et al. \cite{liu2017early} proposed an early active learning algorithm with a pairwise constraint to select the most representative samples for labeling. 
Roy et al. \cite{roy2018exploiting} presented a pairwise training subset selection framework to minimize human annotation effort.
Liu et al. \cite{liu2019deep} designed a deep reinforcement active learning model to minimize human effort in the training stage. 
Hu et al. \cite{hu2021cluster} proposed a purified clustering based active learning framework to learn a discriminative re-id model.
Compared with unsupervised learning and semi-supervised re-id methods, active learning re-id methods iteratively label a handful of pairwise data to improve the re-id performance.
However, these approaches cannot accurately select the critical pairwise data which really contribute to the discriminative feature learning.
In this work, we propose a novel active learning framework to mine and label the critical pairs, i.e. support pairs. Moreover, we propose a constrained clustering algorithm and a hybrid learning loss, to effectively exploit the complementary information of both unlabeled samples and labeled support pairs.


\begin{figure*}[ht]
    \centering
    \scalebox{0.9}{
    \includegraphics{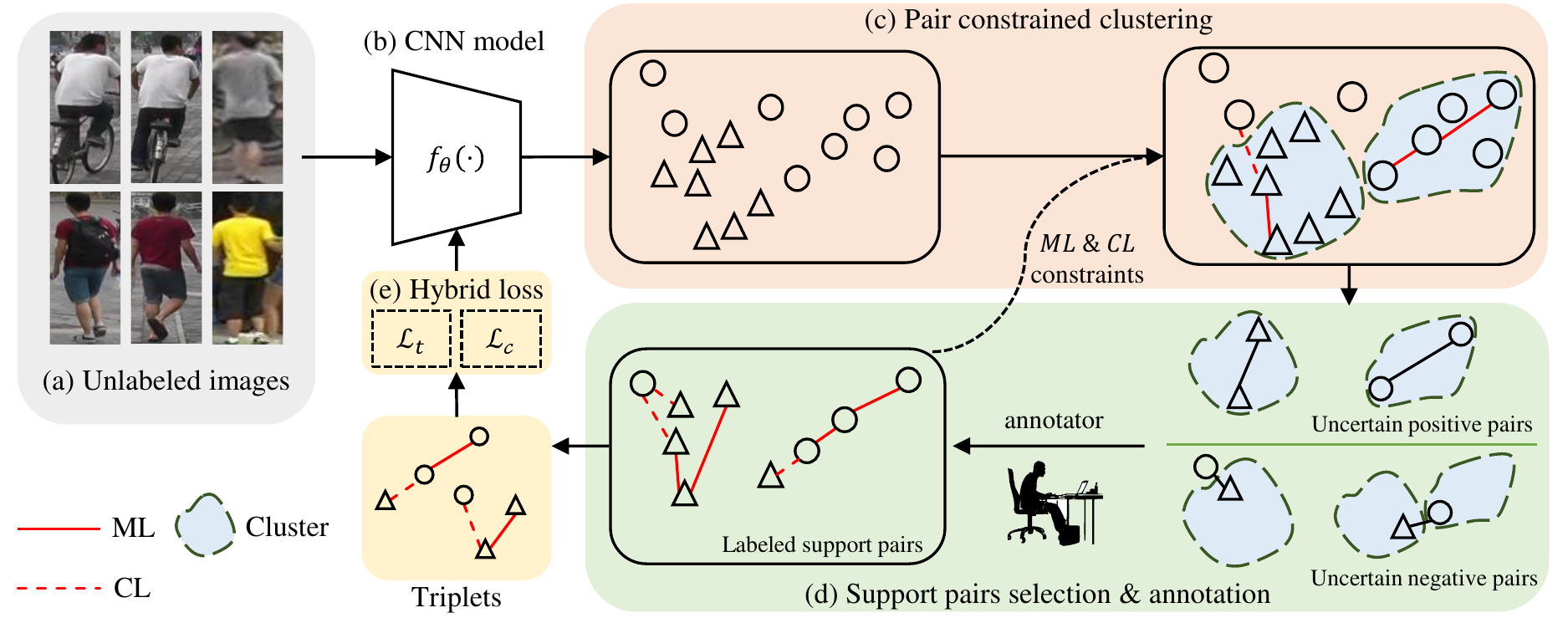}
    }
    \caption{An overview of the proposed Support Pair Active Learning (SPAL) person re-identification model. The SPAL takes as input (a) unlabeled images without any person ID labeling. The objective is to derive (b) a person re-id feature representation model by active learning. To this end, the SPAL adopts (c) a pair constrained clustering algorithm to group both unlabeled and labeled data, and select (d) a set of \textit{support pairs} for annotations by a dual certainty selection strategy. The labeled support pairs are adopted to derive Must-Link(ML) and Cannot-Link(CL) sets, which guides the constrained clustering. After the constrained clustering, (e) a hybrid learning loss is designed to optimize the CNN model iteratively. Best viewed in colour.}
    \label{fig:framework}
\end{figure*}

\section{Method Formulation}

\subsection{Overview}
Given $n$ unlabeled images $\mathcal{I}=\{I_1, I_2, ..., I_n\}$, and a labeling budget $B$, 
active learning re-id methods manage to select a subset of pairs from $\mathcal{I}$ for manual labeling at most $B$, to learn a discriminative feature representation model.
It is critical to select a optimized subset which is the most beneficial for model learning.

To this end, we propose a Support Pair Active Learning (SPAL) framework to mine the critical pairs, termed \textit{support pairs}.
An overview of our proposed method is illustrated in Fig.~\ref{fig:framework}. 
Starting from existing unlabeled person images, the SPAL learns a re-id CNN model which is employed to extract feature embedding $\mathcal{X}=\{x_1, x_2, ..., x_n\}$ of all images.
And then we adopt a clustering algorithm to cluster the unlabeled data and generate pseudo labels.
Although the clustering algorithm inevitably generates incorrect pseudo labels, the clustering result reveals the distribution structure of data.
This is helpful for the critical pairs selection. 
Based on the clustering result, we design a dual uncertainty selection strategy to select the most informative pairwise samples, named support pairs, for human labeling.
For fully utilizing the labeled support pairs, a constrained clustering algorithm is adopted to explore a reliable clustering for all data.
Additionally, we propose a hybrid loss which employs a unsupervised contrastive loss and a supervised support pair loss to train the re-id model effectively.
The overall pipeline of SPAL optimizes the re-id model in an iterative way.

\subsection{Pair Constrained Clustering}
\label{pcc}
Given the extracted feature $\mathcal{X}$ of all instances and the constraints of labeled support pairs, the constrained clustering algorithm aims to generate a reliable clustering while meeting the constrained relations.
We will discuss how to select the support pairs in Sec.\ref{dus}.
Depending on the labels of support pairs, we obtain a set of Must-Link ($ML$) constraints and a set of Cannot-Link ($CL$) constraints. 
We then adopt a Constrained-DBSCAN algorithm \cite{zhao2012effective} to cluster the unlabeled instances by satisfying two kinds of pairwise constraints: $ML$ and $CL$.

Similar with DBSCAN \cite{ester1996density}, Constrained-DBSCAN can discover the clusters of arbitrary shape with noise.
It also requires two parameters: the maximum distance between neighbors $eps$ and the minimal number of neighbors for a core point $minpts$. 
We denote $N_{eps}(p)$ as the $eps$-neighborhood of a point $p$, which is defined by $N_{eps}(p)=\{q\in\mathcal{X}|d(p, q)\leq eps\}$, where $d(\cdot, \cdot)$ denotes the distance between two instances measured by Jaccard distance with $k$-reciprocal nearest neighbors \cite{zhong2017re}. 
The points are divided into three classes: if $|N_{eps}(p)|\geq minpts$, the point $p$ is a core point; if $|N_{eps}(p)|<minpts$ but $p$ in the $eps$-neighborhood of core points, the point $p$ is a border point and the remaining points are outliers which do not belong to any clusters.

Constrained-DBSCAN starts with an arbitrary core point $p$ to expand a cluster by retrieving all points density-reachable from $p$ with respect to $eps$ and $minpts$.
Similar with DBSCAN, Constrained-DBSCAN maintain a seed set to carry out the clustering procedure.
However, there are two main differences in Constrained-DBSCAN algorithm:
(1) must-link constraints force the seed point $q$ and the points involved in the $ML$ set of $q$ adding into the current cluster and seed set.
(2) cannot-link constraints forbid the points involved in the $CL$ set of $q$ adding into the current cluster.
Note that, this way ensures that all pairwise constraints can be satisfied during the clustering.

\paragraph{Discussion.} 
During the clustering process, the impact of the pairwise constraints is reflected in two aspects:
(1) Instance relationship maintenance. $ML$ and $CL$ constraints force two instance points grouped into one cluster or not. It explicitly encourages clustering algorithm to maintain this instance-level relationship.
(2) Cluster relationship discovery. Although the constraints are instance-level, the cluster relationship can be implicitly discovered by the clustering structure, because the instances in the pairwise constraints are inevitably grouped into clusters.
By $ML$ and $CL$ constraints, Constrained-DBSCAN can achieve a more reliable clustering.

\begin{algorithm}[tb] 
\caption{SPAL Framework}
\label{alg:SPAL}
\textbf{Input}: Unlabeled dataset $\mathcal{I}$, constrained pair set $ML=\varnothing$ and $CL=\varnothing$, a CNN-based model $f_{\theta}$ pre-trained on the ImageNet.\\
\textbf{Output}: $f_{\theta}$
\begin{algorithmic}[1] 
\STATE Initialize a memory bank with features $\mathcal{X}$ extracted by $f_{\theta}$;
\FOR{$t=1$ to epochs}
    \STATE Pair Constrained Clustering by $\mathcal{X}$, $ML$ and $CL$;
    \STATE Select support pairs $\mathcal{S}$ by Eq.(\ref{eq:intra}) and Eq.(\ref{eq:inter});
    \STATE Label support pairs $\mathcal{S}$ by human experts;
    \STATE Update $ML$ and $CL$ by labeled support pairs $\mathcal{S}$;
    \FOR{each mini-batch $\subset \mathcal{X}$}
        \STATE Compute the hybrid loss by Eq.(\ref{eq:hybrid}) and update the encoder $f_{\theta}$ by back-propagation;
        \STATE Update the memory bank by Eq.(\ref{eq:memory});
    \ENDFOR
\ENDFOR
\STATE \textbf{return} $f_{\theta}$
\end{algorithmic}
\end{algorithm}

\subsection{Dual Uncertainty Selection}
\label{dus}
Given a re-id dataset which contains $n$ unlabeled instances, obviously it costs ${n(n-1)}/{2}$ times to annotate all pairwise relationships. 
To reduce the labeling cost, we propose a Dual Uncertainty Selection Strategy (DUS) to select the most informative pairs for human annotations. 
These critical pairs are named as \textit{support pairs} in the proposed SPAL.
However, a problem naturally comes up: how to discover the support pairs?

We observe that, the relations of most pairs can be predicted accurately by pseudo labels derived from an unsupervised clustering algorithm. 
However, the points that are located on the border of clusters, such as border point in DBSCAN, are inevitably clustered incorrectly.
In other words, the positive or negative relation of any two border points is the most uncertainty.
Inspired by this, we aim to mine two kinds of \textit{support pairs}: uncertain positive pairs $s_p$ and uncertain negative pairs $s_n$ on an clustering structure at each clustering epoch. 
Specifically, uncertain positive pairs refer to the pairs possibly has a false positive relation, 
while uncertain negative pairs refer to the pairs possibly has a false negative relation.

\paragraph{Uncertain positive pairs selection.}
For effectively mining uncertain positive pairs, we aim to select $s_p$ within each cluster.
The samples grouped in the same cluster are assumed to belong to the same ID.
That is, the relation of any pair among these samples are assumed to be positive.
However, a cluster generated by clustering algorithm inevitably contains false positive pairs.
It is easy to see that, within each cluster, the more dissimilar the pairwise samples are, the more likely they are the false positive. 

Formally, given ${K_t}$ clusters $\mathcal\{C_1, C_2, ..., C_{K_t}\}$ at epoch $t$, for each cluster $C_i$, 
we only select one uncertain positive support pair $s^i_{p}$ as follow:
\begin{align}
\label{eq:intra}
    s^i_{p} = \arg\max_{x_a^i, x_b^i \in C_i} d(x_a^i, x_b^i)
\end{align}
where $d(\cdot, \cdot)$ denotes the Euclidean distance between two samples. 
The pair $s^i_{p}$ is selected as the uncertain positive support pair, because of its the highest uncertainty of a positive relation within cluster $C_i$.

\paragraph{Uncertain negative pairs selection.}
For effectively mining uncertain negative pairs, we aim to select $s_n$ across clusters.
The samples divided to different clusters are assumed to belong to different IDs.
That is, the relation of any pairwise samples across clusters are assumed to be negative.
However, these inter-cluster pairs inevitably contains false negative pairs.
Analogously, it is also easy to see that, the more similar the pairwise inter-cluster samples are, the more likely they are the false negative. 

Formally, given ${K_t}$ clusters $\mathcal{C}=\{C_1, C_2, ..., C_{K_t}\}$ at epoch $t$, for each cluster $C_i$, 
we only select one uncertain negative support pair $s^i_{n}$ as follow:
\begin{align}
\label{eq:inter}
    s^i_{n} = \arg\min_{x_a^i\in C_i, x_b^j\in C_j} d(x_a^i, x_b^j)
\end{align}
where $d(\cdot, \cdot)$ denotes the Euclidean distance between two samples, 
and $C_i \neq C_j$.
The pair $s^i_{n}$ is selected as the uncertain negative support pair, because of its the highest uncertainty of a negative relation for cluster $C_i$.

After the support pairs selection, true positive or true negative relations are required to label by human experts for eliminating the uncertainty.
Like \cite{roy2018exploiting}, a propagation mechanism of positive annotations through the transitive closures is also adopted to save the labeling cost.
After the labeling, the true positive pairs are assigned into $ML$ set, correspondingly the true negative pairs are assigned into $CL$ set.
As described in Sec.\ref{pcc}, in next round of constrained clustering, the instances with the positive constraints in $ML$ are forced into the same cluster and the instances with the negative constraints in $CL$ are forced to different clusters.

\paragraph{Discussion.}
(1) Different from existing active selection strategies \cite{liu2019deep,hu2021cluster}, the proposed DUS strategy only exploits a few of support pairs which are derived from border points located on the edge of clusters, in order to avoid selecting the unnecessary pairs for annotation.
By measuring the intra-cluster uncertainty and the inter-cluster uncertainty of pairs, the most informative pairs are selected for human labeling. 
The support pairs are fewer, but are more critical for model learning.
(2) The DUS strategy selects support pairs based on the border points derived by the clustering algorithm, while the relations generated from labeled support pairs also spread implicitly via the constrained clustering described in Sec.\ref{pcc}.
That is, the active selection strategy and semi-supervised clustering are unified and complementary in our proposed SPAL framework.

\subsection{Hybrid Loss Learning}
Given the unlabeled training dataset, we firstly adopt a semi-supervised clustering algorithm to group the data into clusters. We update the model according to the results of clustering and the pairwise constraints from human expert.
As the number of clusters and outliers may change with the alternate clustering strategy, the class prototypes are built in a non-parametric and dynamic manner. 
Following \cite{ge2020selfpaced}, we maintain a memory bank $\mathcal{M}$ to store an average approximation to all features $\{\overline{x}_1, \overline{x}_2, ..., \overline{x}_n\}$, and the memory is updated with the encoded features in a mini-batch by:
\begin{equation}
\label{eq:memory}
    \overline{x}_i \leftarrow \alpha\overline{x}_i+(1-\alpha)x_i
\end{equation}
where $\alpha$ is the momentum coefficient for updating features.

Based on the memory bank, we first adopt instance-to-centroid contrastive loss to push the instance close to the centroid of its cluster and far from other clusters. 
The formula is as follows:
\begin{align}
\label{eq:Lc}
    \mathcal{L}_c=-\log\frac{\exp(x\cdot c_+/\tau)}{\sum_{i=1}^N{\exp(x\cdot c_i/\tau)}}
\end{align}
where $N$ is the number of clusters and outliers, $c_i$ is the cluster centroid corresponding to pseudo label $i$ or outliers itself,
and $\tau$ is temperature factor.

To effectively utilize the labeled support pairs, we propose a triplet loss based on support pairs constraints.
This loss encourages an anchor sample closer to its hard must-link instance and away from its hard cannot-link instance. 
Different from Eq.(\ref{eq:Lc}), the support pair loss is instance-to-instance, which makes the clusters more independent and compact. 
Consequently, the support pair constraints loss is formulated as below:
\begin{equation}
\label{eq:support}
\mathcal{L}_t=-\frac{1}{M}\sum_{i=1}^M{\max(0, d(x_i, x_{ml})-d(x_i,x_{cl})+m)}  
\end{equation}
where $x_i$ is an anchor in a mini-batch, $x_{ml}$ is the hardest positive in $ML$ and $x_{cl}$ is the hardest negative in $CL$ corresponding to $x_i$, and $M$ is the total number of valid triplets. 

Finally, we train the model by combining the unsupervised contrastive loss $\mathcal{L}_c$ and the supervised support-pair triplet loss $\mathcal{L}_t$ as follow:
\begin{equation}
\label{eq:hybrid}
\mathcal{L}_{hybrid}=\mathcal{L}_c+\lambda \mathcal{L}_t
\end{equation}
where $\lambda$ is the hyper-parameter for balancing the two loss functions.
As such, the hybrid loss function optimizes the re-id model by utilizing both unlabeled data and labeled data.

\section{Experiments}

\subsection{Datasets and Evaluation Protocol}

\paragraph{Datasets.} We evaluate our proposed method on four large-scale person re-identification datasets: {\it Market-1501}\cite{zheng2015scalable}, {\it DukeMTMC-ReID}\cite{zheng2017unlabeled}, {\it MSMT17}\cite{wei2018person}, and {\it LaST}\cite{shu2021large}.
We adopted the standard person re-id setting on training/test ID split and the test protocols (Table \ref{tab:dataset_stats}).





\begin{table}[h]
	\begin{center}
	\footnotesize
	\setlength{\tabcolsep}{0.2cm}
	\begin{tabular}{c||c|c||c|c}
	    \toprule
		\multirow{2}{*}{Dataset} &\multicolumn{2}{c||}{Training} &\multicolumn{2}{c}{Test} \\
		\cline{2-5}
		& {\# Image} & {\# ID } & {\# Image} & {\# ID}
        \\ \hline \hline %
		Market-1501		& 12,936 & 751 & 19,281  & 751 		\\
		DukeMTMC-ReID	& 16,522 & 702 & 19,889 & 1,110		\\
		MSMT17		    & 32,621 & 1,041 & 93,820 & 3,060 	\\
		LaST		    & 71,248 & 5,000 & 135,529 & 5,806 	\\
	    \bottomrule
	\end{tabular}
	\caption{Dataset statistics.}
	\label{tab:dataset_stats}
	\end{center}
\end{table}


\paragraph{Evaluation Protocol.}Mean average precision (mAP) and cumulative matching characteristic (CMC) are adopted to measure the method's performance.

\paragraph{Labeling Budget.}
In this work, labeling budget is computed via the number of labeled pairs.
Given the training image number $n$, the maximum budget is set to $n(n-1)/2$.

\subsection{Implementation Details}
We adopt DBSCAN + Contrastive-Learning as our baseline method, i.e. unsupervised learning framework. 
This is similar with SPCL\cite{ge2020selfpaced} but without the self-paced learning strategy.
For DBSCAN, the maximum distance between neighbors is set to 0.6 and the minimum number of neighbors for a core point is set to 4.
In the baseline method, only $\mathcal{L}_c$ (Eq.(\ref{eq:Lc})) is used for unsupervised learning.

We use an ImageNet pre-trained ResNet-50 as the backbone for the encoder $f_\theta$. 
After pooling-5 layer, we remove subsequent layers and add a 1D BatchNorm layer and an L2-normalization to derive the feature representations.
The input images are resized to 256$\times$128. 
Adam optimiser with a weight decay of 0.0005 is adopted to optimise our model. The initial learning rate is set to 0.00035. We train the model for 50 epochs. 
The temperature $\tau$ is empirically set to 0.05 and the momentum coefficient $\alpha$ is set to 0.2. 
The margin $m$ in Eq.(\ref{eq:support}) is set to 1.0, 
and the balance coefficient $\lambda$ in Eq.(\ref{eq:hybrid}) is set to 1 to balance two loss functions.

\begin{table}[t]
    \begin{center}
        \footnotesize
        \resizebox{1.\linewidth}{!}{
        \begin{tabular}{c|c||c|c|c||c|c|c}
        \toprule
        \multicolumn{2}{c||}{\multirow{2}{*}{Methods}} & \multicolumn{3}{c||}{Market-1501} & \multicolumn{3}{c}{DukeMTMC-ReID}\\
        \cline{3-8}
        \multicolumn{2}{c||}{} & Rank1 & mAP & Budget & Rank1 & mAP & Budget\\
        \hline
        \multirow{7}{*}{UL}
        & BUC & 66.2 & 38.3 & 0 & 47.4 & 27.5 & 0\\
        & ECN & 75.1 & 43.0 & 0 & 63.3 & 40.4 & 0\\
        & MMCL & 80.3 & 45.5 & 0 & 65.2 & 40.2 & 0\\
        & HCT & 80.0 & 56.4 & 0 & 69.6 & 50.7 & 0\\
        & SSG & 80.0 & 58.3 & 0 & 73.0 & 53.4 & 0\\
        & SPCL & 89.7 & 77.5 & 0 & - & - & -\\
        & Baseline & 85.4 & 67.7 & 0 & 79.3 & 62.5 & 0\\
        \hline
        \multirow{4}{*}{SSL}
        & EUG & 49.8 & 22.5 & - & 45.2 & 24.5 & -\\
        & TAUDL & 63.7 & 41.2 & - & 61.7 & 43.5 & -\\
        & UTAL & 69.2 & 46.2 & - & 62.3 & 44.6 & -\\
        & SSG++ & 86.2 & 68.7 & - & 76.0 & 60.3 & - \\
        \hline
        \multirow{7}{*}{AL}
        & TMA & 47.9 & 22.3 & - & - & - & -\\
        & QIU & 67.8 & 45.0 & - & 56.8 & 36.8 & - \\
        & QBC & 68.4 & 46.3 & - & 61.1 & 40.8 & -\\
        & GD & 71.4 & 49.3 & - & 53.5 & 33.6 & - \\
        & DRAL & 84.2 & 66.3 & 10n & 74.3 & 56.0 & 10n\\
        & MASS & 93.5 & 81.7 & 5n & 86.1 & 72.9 & 5n\\
        & SPAL(ours) & \textbf{94.0} & \textbf{83.1} & \textbf{2n} & \textbf{86.4} & \textbf{73.3} & \textbf{1.8n}\\
        \hline
        SL & Oracle & 94.0 & 84.6 & n(n-1)/2 & 87.2 & 75.3 & n(n-1)/2\\
        \bottomrule
        \end{tabular}
        }
    \end{center}
    \caption{Comparison with state-of-the-art methods on Market-1501 and DukeMTMC-reID.}
    \label{tab:market_duke}
\end{table}

\begin{table}[t]
    \begin{center}
        \footnotesize
        \resizebox{1.\linewidth}{!}{
        \begin{tabular}{c|c||c|c|c||c|c|c}
        \toprule
        \multicolumn{2}{c||}{\multirow{2}{*}{Methods}} & \multicolumn{3}{c||}{MSMT17} & \multicolumn{3}{c}{LaST}\\
        \cline{3-8}
        \multicolumn{2}{c||}{} & Rank1 & mAP & Budget & Rank1 & mAP & Budget\\
        \hline
        \multirow{5}{*}{UL}
        & ECN & 25.3 & 8.5 & 0 & - & - & -\\
        & MMCL & 35.4 & 11.2 & 0 & - & - & -\\
        & SSG & 32.2 & 13.3 & 0 & - & - & -\\
        & SPCL & 53.7 & 26.8 & 0 & 44.9 & 10.1 & 0\\
        & Baseline & 45.5 & 20.0 & 0 & 45.1 & 9.9 & 0\\
        \hline
        \multirow{3}{*}{SSL}
        & TAUDL & 28.4 & 12.5 & - & - & - & -\\
        & UTAL & 31.4 & 13.1 & - & - & - & -\\
        & SSG++ & 41.6 & 18.3 & - & - & - & -\\
        \hline
        \multirow{5}{*}{AL}
        & QIU & 31.3 & 21.5 & - & - & - & -\\
        & QBC & 39.4 & 24.9 & - & - & - & -\\
        & GD & 42.5 & 25.4 & - & - & - & -\\
        & MASS & 54.1 & 30.0 & 5n & - & - & -\\
        & SPAL(ours) & \textbf{69.0} & \textbf{41.0} & \textbf{1.8n} & \textbf{53.3} & \textbf{13.9} & \textbf{1.4n}\\
        \hline
        SL & Oracle & 72.3 & 46.4 & n(n-1)/2 & 59.3 & 19.1 & n(n-1)/2\\
        \bottomrule
        \end{tabular}
        }
    \end{center}
    \caption{Comparison with state-of-the-art methods on MSMT17 and LaST.}
    \label{tab:msmt_last}
\end{table}

\subsection{Compared with State-of-the-art Methods}
The proposed SPAL is compared with state-of-the-art methods including 
unsupervised learning (UL) methods: 
BUC\cite{lin2019aBottom}, ECN\cite{zhong2019invariance}, MMCL\cite{wang2020unsupervised}, HCT\cite{zeng2020hierarchical}, SSG\cite{fu2019self}, SPCL\cite{ge2020selfpaced}, 
semi-supervised learning(SSL) methods: 
EUG\cite{wu2018exploit}, TAUDL\cite{li2018unsupervised}, UTAL\cite{li2019unsupervised}, SSG++\cite{fu2019self}, 
and active learning(AL) methods: GD\cite{ebert2012ralf}, QIU\cite{lewis1994sequential}, QBC\cite{abe1998query}, DRAL\cite{liu2019deep}, MASS\cite{hu2021cluster}.
The "Oracle" experiment is a supervised setting, which utilizes all the pairwise relationship of dataset with $n$ images for training with the hybrid learning loss.
In Table.\ref{tab:market_duke} and \ref{tab:msmt_last}, we illustrate Rank1 and mAP measurements on Market-1501, DukeMTMC-reID, MSMT17 and LaST benchmarks.

\paragraph{Comparison with unsupervised learning methods.}
We compare SPAL with UL methods including the pure unsupervised methods and the unsupervised domain adaption methods.
Although the labeling cost is zero in UL, the performance is not satisfactory.
With a low labeling cost, the proposed SPAL surpasses existing unsupervised learning methods on four datasets.
For example, the mAP margin by SPAL over SPCL is 5.6\% on Market and 14.2\% on MSMT17.
These results verify that the performance of UL methods is limited without human labeling, and the SPAL is capable of significantly improving performance with relatively few annotations.

\paragraph{Comparison with semi-supervised learning methods.}
The semi-supervised learning methods follow an impractical setting, that need annotate all IDs to select supervised data.
The proposed SPAL outperforms all SSL competitors by incrementally labeling data without any ID information.
It reveals that selecting the critical pairwise data for labeling benefits the performance of person re-identification.

\paragraph{Comparison with active learning methods.}
We compare the proposed SPAL with the state-of-the-art active learning based approaches and three active learning strategies.
The proposed SPAL method outperforms all these methods with significant margins on the performance and the annotation cost especially on large-scale benchmarks.
Specially, on larger dataset MSMT17, compared with the state-of-the-art method MASS, SPAL improves the performance of mAP and Rank1 by 11.0\% and 14.9\%  under a fewer annotation cost of 1.8n, revealing the effectiveness and efficiency of the SPAL.

\paragraph{Comparison with supervised learning methods.}
To measure the gap between active learning and supervised learning model, we compare the proposed SPAL model with the supervised model with the fully labeled pairs (i.e. the upper bound of SPAL, denoted as “Oracle”).
Table \ref{tab:market_duke} and \ref{tab:msmt_last} show that, with much fewer annotations, the SPAL achieves a competitive performance compared with the supervised learning method.

\subsection{Ablation Studies}
\paragraph{Components analysis.}
We analyse the effectiveness of the proposed DUS strategy and the hybrid learning loss in Table.\ref{tab:ablation_study_market} demonstrate that, 
compared with the unsupervised learning model (shown in the top row), 
uncertain positive pairs $s_p$ and uncertain negative pairs $s_n$ selected by DUS are both conducive to model learning.
Additionally, Table.\ref{tab:ablation_study_market} shows a distinct performance benefit from the hybrid learning loss, e.g. a Rank-1 boost of 2.0\% and a mAP boost of 2.9\% on Market-1501.
This validates the importance of jointly learning from both unlabeled data and labeled data and the effectiveness the hybrid learning loss formulation.


\paragraph{Different budgets analysis.}
We performed an analysis on the model performance with different labeling budgets and conducted a controlled evaluation on MSMT17.
During the training epochs, when the labeling budget runs out, the support pair selection operation stops and the model is subsequently trained on the constrained clustering results.
Along with the increasing budget, the model improves significantly.
Moreover, with a fewer budget, the SPAL outperforms the state-of-the-art method MASS.
We set the labeling budget to $2n$ for all the experiments.
The actual labeling cost is reported if the cost is less than the budget.

\begin{figure}[!h]
    \begin{center}
    \includegraphics[width=0.9\linewidth]{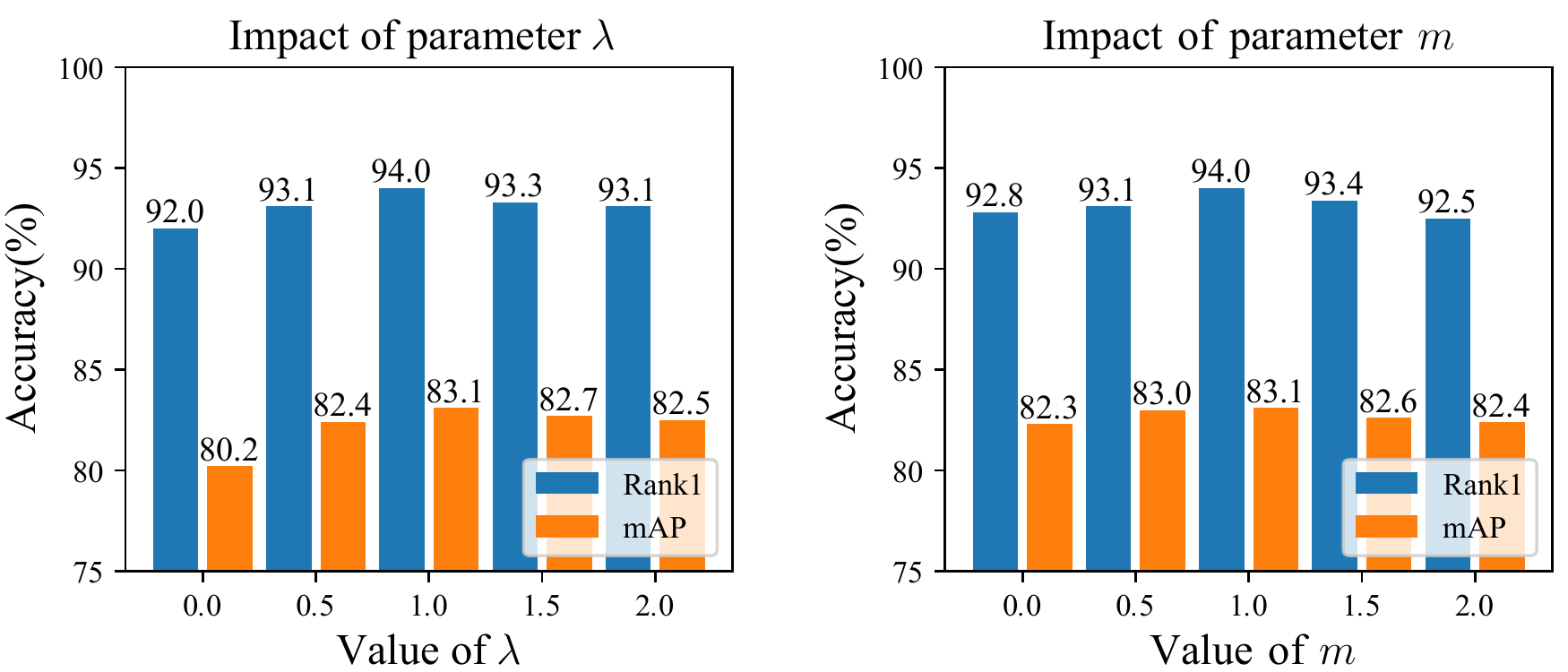}
    \end{center}
    \caption{Parameter analysis on $\lambda$ and $m$ on Markter-1501.}
    \label{fig:lambda_m}
    \vspace{-3mm}
\end{figure}

\paragraph{Parameter sensitive analysis.} We analyse the impact of the hyper-parameter $\lambda$ in Eq.(\ref{eq:hybrid}) and $m$ in Eq.(\ref{eq:support}). 
As shown in Fig.\ref{fig:lambda_m}, $\lambda$ is not sensitive with a wide satisfactory range.
This suggests a stable model learning procedure.
Specially, when $\lambda$ equals to 0, it can be considered to train the model only with contrastive loss function.
On the other hand, the SPAL achieves the similar performance when $m$ changes, which indicates that the proposed method is also not sensitive to the hyper-parameter $m$. 
In this work, We set $\lambda=1.0$ and $m=1.0$ for all the experiments on four person re-id datasets.


\begin{table}[t]
    \begin{center}
    \footnotesize
    \begin{tabular}{cccc|c|c|c|c}
    \toprule
    \multicolumn{4}{c|}{Settings} & \multicolumn{4}{c}{Market-1501}\\
    \hline
    $s_{p}$ & $s_{n}$ & $\mathcal{L}_c$ & $\mathcal{L}_t$ & mAP & Rank1 & Rank5 & Rank10\\
    \hline
    & & \checkmark & & 67.7 & 85.4 & 93.2 & 95.1\\
    \hline
    \checkmark & & \checkmark & & 77.4 & 90.4 & 96.7 & 98.2\\
    & \checkmark & \checkmark & & 76.3 & 90.4 & 95.9 & 97.6\\
    \checkmark & \checkmark & \checkmark & & 80.2 & 92.0 & 96.7 & 98.0\\
    \hline
    \checkmark & & \checkmark& \checkmark & 79.6 & 91.6 & 96.6 & 97.9 \\
    & \checkmark & \checkmark & \checkmark & 78.0 & 91.0 & 96.6 & 98.0\\
    \checkmark & \checkmark & \checkmark  & \checkmark & 83.1 & 94.0 & 97.8 & 98.5 \\
    \bottomrule
    \end{tabular}
    \caption{Effect of DUS strategy and hybrid learning loss on Market1501.}
    \label{tab:ablation_study_market}
    \end{center}
\end{table}

\begin{table}[t]
\begin{center}
    \footnotesize
    \begin{tabular}{c|c|c|c|c|c}
    \toprule
    \multirow{2}{*}{Method} & \multicolumn{5}{c}{MSMT17} \\
    \cline{2-6}
    \multicolumn{1}{c|}{} & mAP & Rank1 & Rank5 & Rank10 & Budget\\
    \hline
        MASS & 30.0 & 54.1 & 65.4 & 70.4 & 5n\\
    \hline
    \multirow{4}{*}{SPAL}
        & 20.0 & 45.5 & 57.2 & 62.8 & 0\\
        & 32.0 & 60.4 & 72.3 & 76.8 & n\\
        & 40.3 & 68.3 & 79.8 & 83.8 & 1.5n\\
        & 41.0 & 69.0 & 80.7 & 84.8 & 1.8n\\
    \hline
    Oracle & 46.6 & 72.3 & 84.7 & 88.7 & n(n-1)/2\\
    \bottomrule
    \end{tabular}
    \caption{Model performance analysis on different budgets on MSMT17.}
    \label{tab:budget}
\end{center}
\end{table}

\section{Conclusion}

In this work, we propose a Support Pair Active Learning (SPAL) framework for large-scale person re-identification to reduce the labeling cost while achieving competitive re-identification performance. 
Specifically, a dual uncertainty selection strategy is proposed to incrementally discover support pairs and require human annotations.
Moreover, we propose a constrained clustering algorithm and a hybrid learning loss, to exploit the complementary information of both unlabeled samples and labeled support pairs.
Extensive comparative experiments demonstrate the two-fold advantages of SPAL: high re-id performance and low labeling cost on four large-scale person re-id benchmarks.


\section*{Acknowledgment}
This work is supported by National Natural Science Foundation of China (Project No. 62076132) 
and Natural Science Foundation of Jiangsu (Project No. BK20211194).

\bibliographystyle{named}
\bibliography{ijcai22}

\end{document}